\documentclass[runningheads]{llncs}

\usepackage{eccv}

\usepackage{eccvabbrv}

\usepackage{graphicx}
\usepackage{booktabs}
\usepackage{algorithm}
\usepackage{algpseudocode}

\usepackage{multirow}

\usepackage[accsupp]{axessibility}  

\usepackage{hyperref}

\usepackage{orcidlink}

\begin{document}

\title{SkipGS: Post-Densification Backward Skipping for Efficient 3DGS Training} 

\titlerunning{SkipGS}
\author{Jingxing Li\orcidlink{0009-0005-2503-8843} \and
Yongjae Lee\orcidlink{0000-0003-1692-2117} \and
Deliang Fan\orcidlink{0000-0002-7989-6297}}

\authorrunning{Jingxing Li et al.}

\institute{Arizona State University, Tempe, AZ 85281, USA
\email{\{jingxing,ylee298,dfan\}@asu.edu}}
\maketitle

\begin{figure*}[t]
  \centering
  \begin{minipage}{\linewidth}
    \centering
    \setlength{\tabcolsep}{3pt}
    \begin{tabular}{cccc}
      \multicolumn{2}{c}{\small Vanilla 3DGS~\cite{kerbl3Dgaussians}} &
      \multicolumn{2}{c}{\small FastGS~\cite{feng2024fastgs}} \\
      {\small Baseline} & {\small +SkipGS (Ours)} &
      {\small Baseline} & {\small +SkipGS (Ours)} \\
      \includegraphics[width=0.23\linewidth]{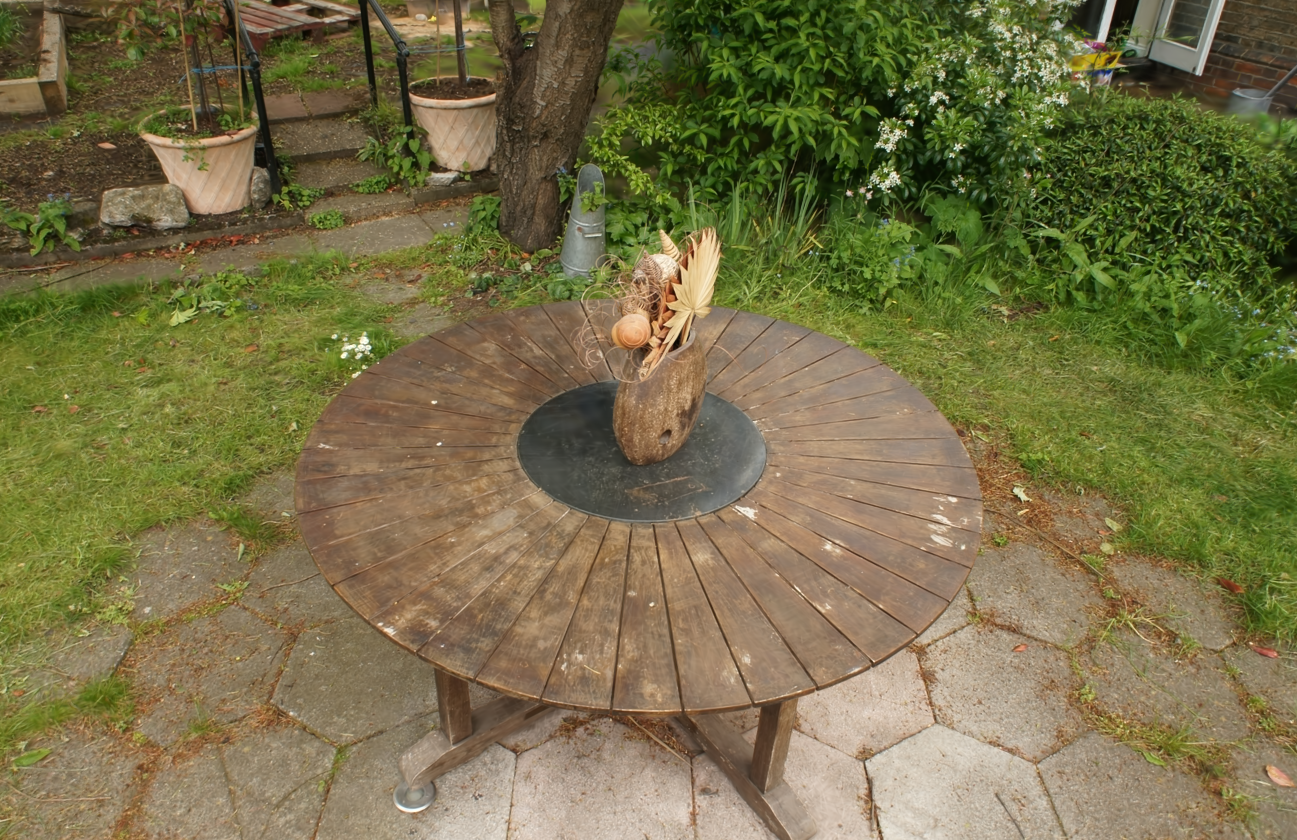} &
      \includegraphics[width=0.23\linewidth]{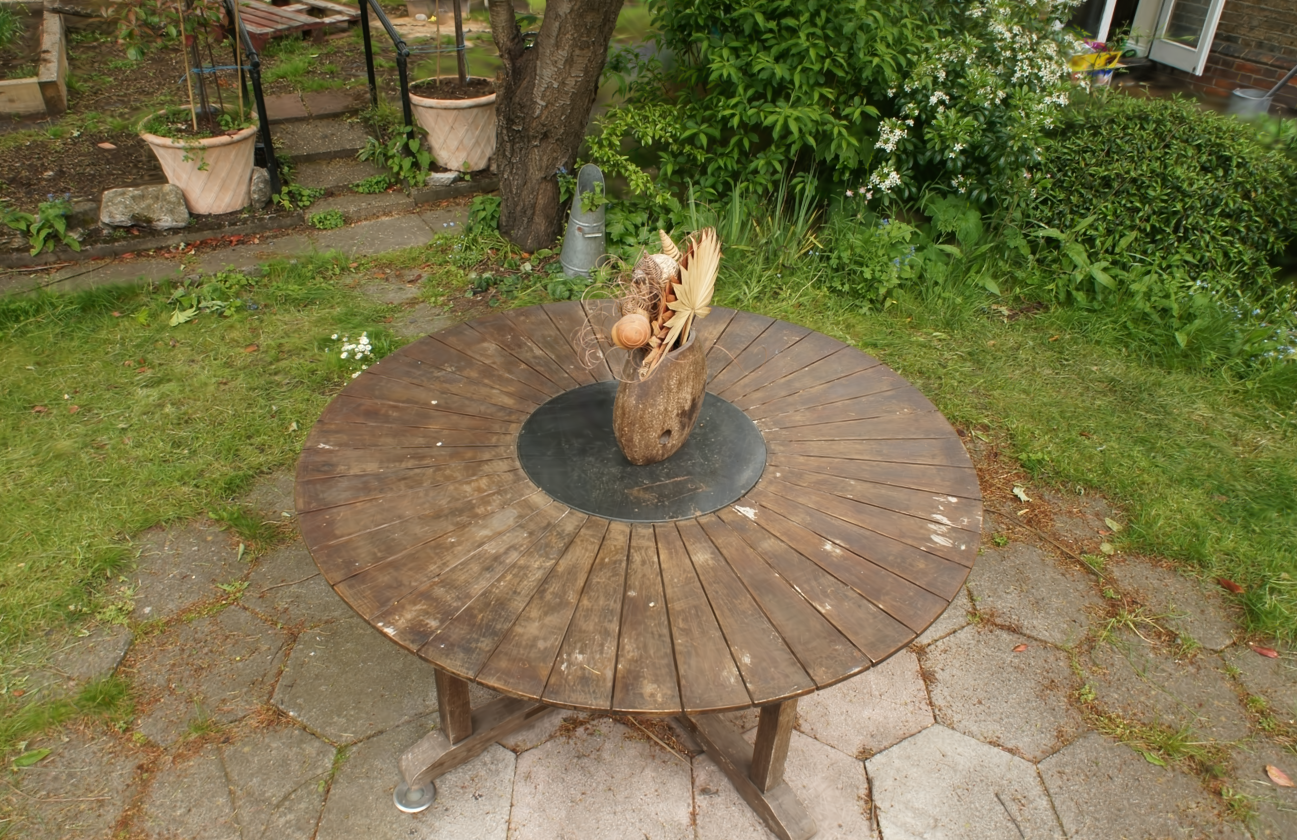} &
      \includegraphics[width=0.23\linewidth]{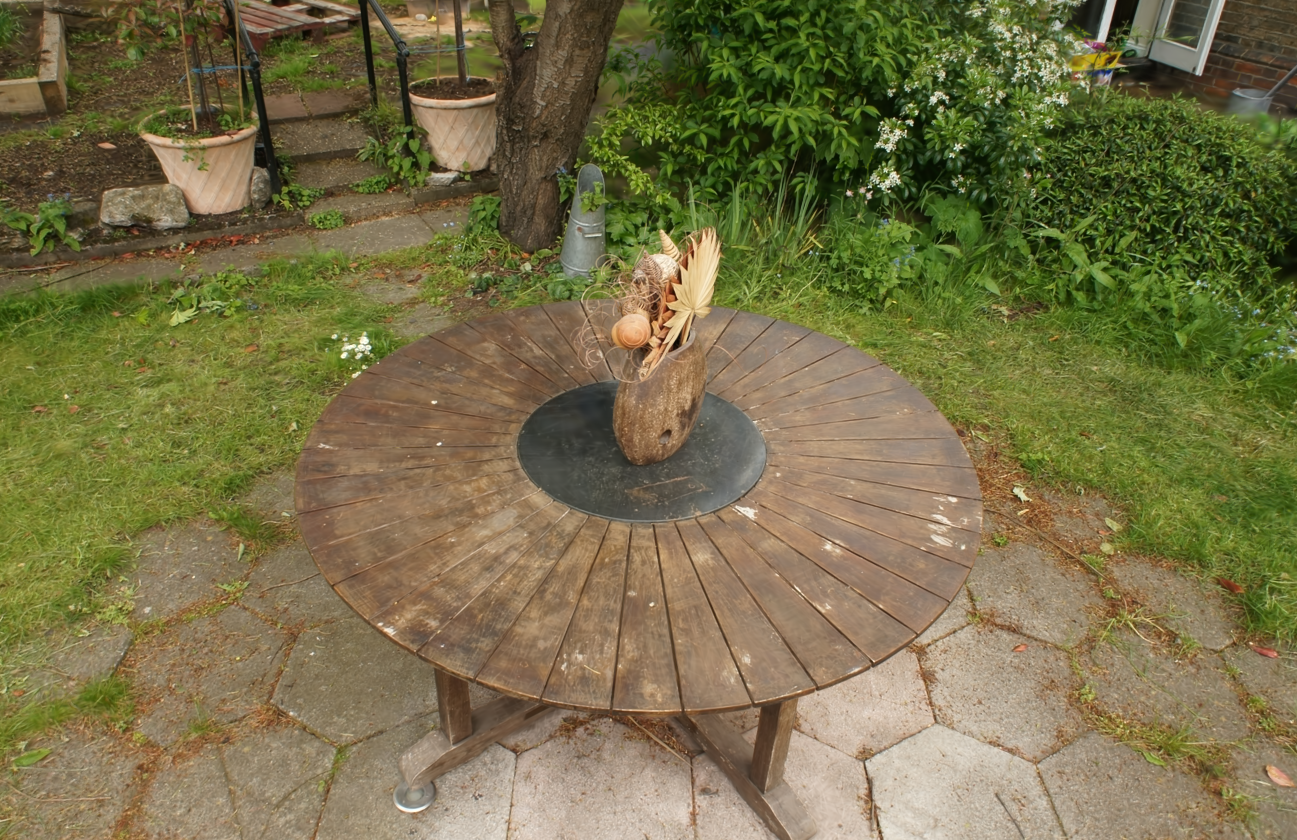} &
      \includegraphics[width=0.23\linewidth]{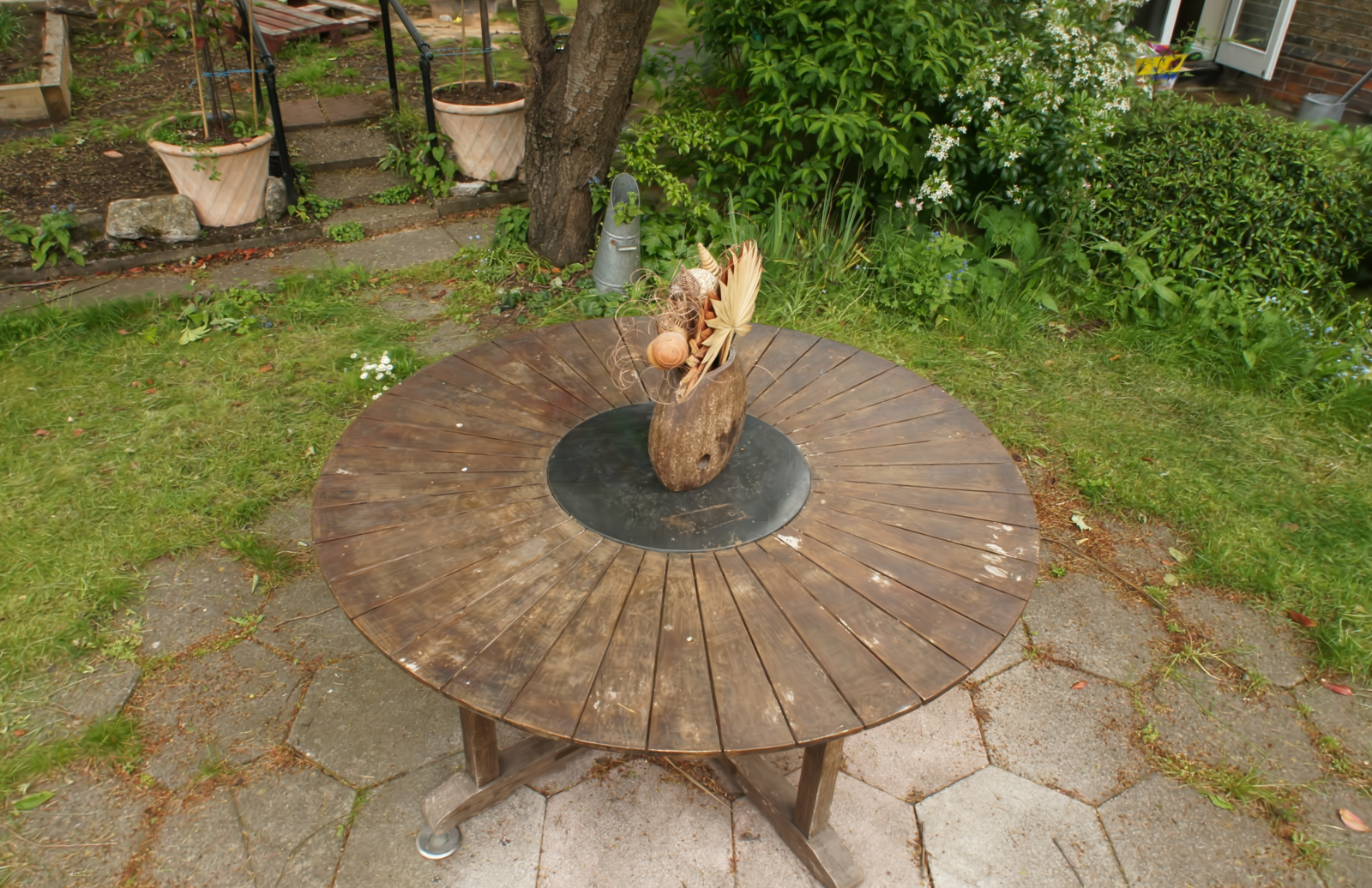} \\
    \end{tabular}
  \end{minipage}

  \vspace{6pt}

  \begin{minipage}{\linewidth}
    \centering
    \includegraphics[width=0.75\linewidth]{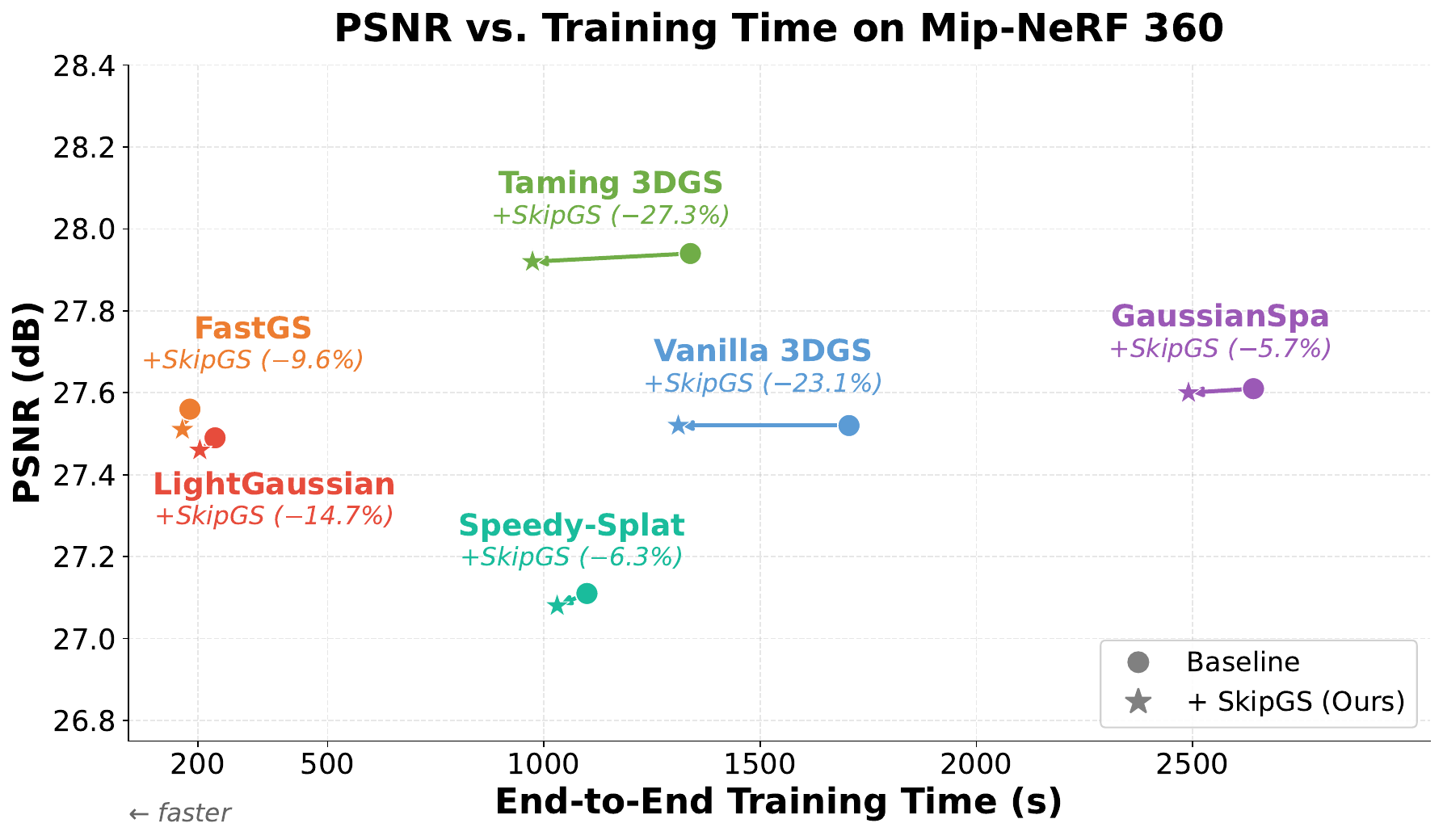}
  \end{minipage}
  \caption{\textbf{SkipGS accelerates diverse 3DGS pipelines with negligible quality loss.}
  \textit{Top}: Qualitative comparison on the \textit{garden} scene (Mip-NeRF\,360~\cite{barron2022mipnerf360}). Baseline and +SkipGS renderings are visually indistinguishable.
  \textit{Bottom}: PSNR vs.\ training time for all six baselines.}
  \label{fig:teaser}
\end{figure*}

\begin{abstract}
3D Gaussian Splatting (3DGS) achieves real-time novel-view synthesis by optimizing millions of anisotropic Gaussians, yet its training remains expensive, with the backward pass dominating runtime in the post-densification refinement phase.
We observe substantial update redundancy in this phase: many sampled views have near-plateaued losses and provide diminishing gradient benefits, but standard training still runs full backpropagation.
We propose \emph{SkipGS} with a novel view-adaptive backward gating mechanism for efficient post-densification training.
SkipGS always performs the forward pass to update per-view loss statistics, and selectively skips backward passes when the sampled view's loss is consistent with its recent per-view baseline, while enforcing a minimum backward budget for stable optimization.
On Mip-NeRF~360, compared to 3DGS, SkipGS reduces end-to-end training time by \textbf{23.1\%}, driven by a \textbf{42.0\%} reduction in post-densification time, with comparable reconstruction quality.
Because it only changes when to backpropagate without modifying the renderer, representation, or loss, SkipGS is plug-and-play and compatible with other complementary efficiency strategies, enabling additive speedups. Code is available at \href{https://github.com/ASU-ESIC-FAN-Lab/SkipGS}{\textcolor{magenta}{https://github.com/ASU-ESIC-FAN-Lab/SkipGS}}.
\keywords{3D Gaussian Splatting \and Training acceleration \and Novel-view synthesis}
\end{abstract}

\section{Introduction}
\label{sec:intro}

3D Gaussian Splatting (3DGS)~\cite{kerbl3Dgaussians} has become the state-of-the-art (SOTA) baseline for real-time novel-view synthesis by optimizing millions of anisotropic Gaussians with a fully differentiable rasterizer.
Despite its rendering efficiency, training remains expensive: modern 3DGS pipelines typically require tens of thousands of iterations, and the backward pass dominates runtime once the Gaussian set grows to a saturated number of primitives.
This training cost is a practical bottleneck for interactive scene capture, large-scale benchmarking, and downstream pipelines that require repeated 3DGS fitting.

Most existing efforts to accelerate 3DGS training focus on the \emph{primitive dimension}: they reduce the number of Gaussians via pruning/compaction, or constrain Gaussian growth during densification under a resource budget.
While effective, these approaches primarily optimize \emph{how many} primitives are processed per iteration.
 
They leave largely unexplored an orthogonal opportunity arising from a distinctive property of 3DGS training: its two-phase structure induced by densification.

During the early \emph{densification} phase, the Gaussian set expands rapidly and optimization must frequently revisit many views to establish geometry and appearance.
In the subsequent \emph{post-densification} phase, the Gaussian
set becomes fixed and training turns into a long refinement stage.
In this regime, we observe substantial update redundancy: per-view losses stabilize
at different rates, and many sampled views yield diminishing returns, yet standard 3DGS~\cite{kerbl3Dgaussians} still
executes a full backward pass whenever a view is sampled.

\begin{figure*}[t]
  \centering
  \includegraphics[width=0.49\linewidth]{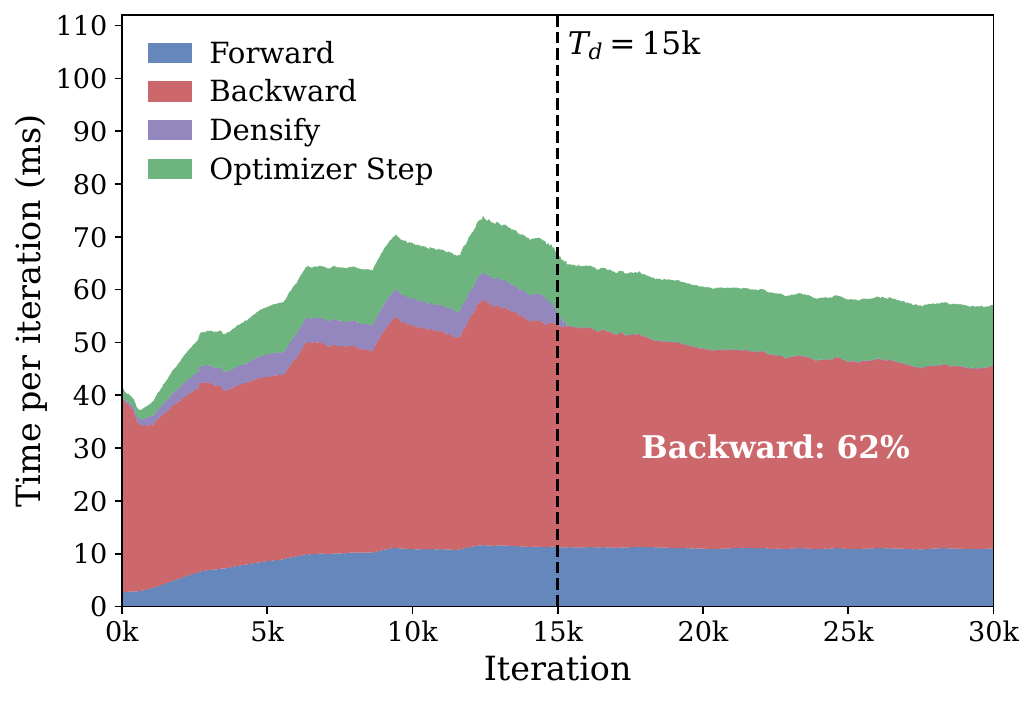}\hfill
  \includegraphics[width=0.49\linewidth]{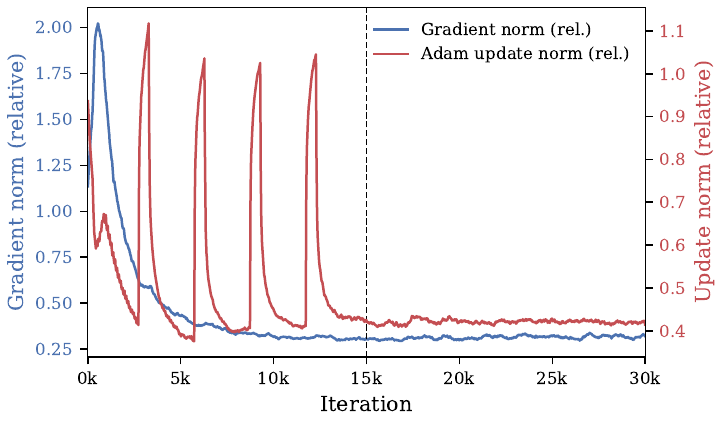}

  \caption{\textbf{Profiling vanilla 3DGS training on the \textit{Kitchen} scene (Mip-NeRF\,360).} \textbf{(a)}~Per-iteration time breakdown: the backward pass dominates (${\sim}62\%$) after densification stops at $T_d{=}15$k, motivating backward-level acceleration. \textbf{(b)}~Average per-Gaussian gradient norms (blue, left axis) decrease ${\sim}2{\times}$ from early to late training and become nearly flat after $T_d$, while average Adam update norms (red, right axis) remain comparatively stable (only ${\sim}1.2{\times}$ reduction overall) due to momentum inertia, suggesting many post-densification updates are weakly informative and can be reduced by selective backpropagation. In (b), both norms are normalized by their respective values at iteration $T_d{=}15$k for cross-quantity comparability.}
  \label{fig:motivation}
\end{figure*}

We quantify this redundancy through profiling analysis of vanilla 3DGS training.
As shown in \cref{fig:motivation}(a), the backward pass dominates per-iteration cost
(${\sim}62\%$) after densification stops, making it the primary target for acceleration.
Meanwhile, \cref{fig:motivation}(b) reveals that per-Gaussian gradient norms decrease
${\sim}2{\times}$ from early to late training and become nearly flat after $T_d$, while
average Adam update norms remain comparatively stable due to momentum inertia, indicating that
many post-densification backward passes produce weakly informative gradients that
contribute little to parameter updates.
Together, these observations suggest that a significant fraction of post-densification
backward passes are redundant: the dominant cost component yields diminishing optimization
benefit.
This motivates a question: can we reduce training time by executing backward
passes only when a sampled view is expected to provide useful optimization signal?

We propose \emph{SkipGS}, a view-adaptive backward gating mechanism for post-densification 3DGS training.
SkipGS retains the standard forward pass to compute the loss and update per-view statistics, but selectively skips backward passes when the sampled view's loss is consistent with its recent per-view baseline.
To ensure stable optimization, we include warmup initialization and enforce a minimum backward budget.
Because our method operates at the level of \emph{when} to backpropagate rather than changing the renderer, Gaussian representation, or loss/optimizer formulation, it is plug-and-play for existing 3DGS variants and compatible with other complementary acceleration directions such as pruning and budgeted densification~\cite{feng2024fastgs,mallick2024taming3dgs,speedysplat}, yielding additive speedups while preserving reconstruction quality.

The major contributions of this work include:
\begin{itemize}
    \item \textbf{Post-densification backward skipping.} We propose \textit{SkipGS}, a view-adaptive backward gating mechanism for the post-densification phase of 3DGS. SkipGS always runs the forward pass to track per-view loss statistics, and skips redundant backward passes when the sampled view's loss is consistent with its recent per-view baseline, while enforcing stability with a minimum backward budget.

  \item \textbf{Speedups with preserved quality.} We show that SkipGS yields substantial wall-clock training speedups in the post-densification phase without compromising reconstruction quality. On Mip-NeRF~360, it reduces end-to-end training time by \textbf{23.1\%}, driven by a \textbf{42.0\%} reduction in post-densification time over 3DGS, while maintaining comparable rendering quality.

  \item \textbf{Orthogonal compatibility.} Because SkipGS operates solely at the backward gating level without modifying the renderer, representation, or loss, it can be applied on top of existing 3DGS variants as a post-densification plug-in. We empirically demonstrate that SkipGS \textit{further accelerates} representative state-of-the-art efficient 3DGS methods, including FastGS~\cite{feng2024fastgs}, Taming 3DGS~\cite{mallick2024taming3dgs}, GaussianSpa~\cite{gaussianspa}, LightGaussian~\cite{lightgaussian}, and Speedy-Splat~\cite{speedysplat}, yielding additional wall-clock savings while maintaining comparable reconstruction quality across three standard benchmarks.

\end{itemize}

\section{Related Work}
\label{sec:related}

\paragraph{3D Gaussian Splatting.}
3D Gaussian Splatting (3DGS)~\cite{kerbl3Dgaussians} represents scenes as anisotropic Gaussians and enables efficient differentiable rendering with a tile-based rasterizer.
Its training pipeline naturally follows a two-phase structure: a densification phase that grows and refines the Gaussian set, followed by a post-densification refinement phase where the Gaussian set is fixed and parameters are fine-tuned.
We leverage this phase transition to reduce redundant computation without changing the representation.

\paragraph{Gaussian compaction.}
A dominant direction improves efficiency by explicitly reducing the number of Gaussians, which lowers rendering cost and can reduce the per-iteration workload in optimization.
FastGS~\cite{feng2024fastgs} integrates training-time pruning with pipeline optimizations and explicitly targets wall-clock training acceleration.
GaussianSpa~\cite{gaussianspa} progressively enforces sparsity (e.g., via opacity- or importance-based criteria) during training to obtain a compact Gaussian set.
LightGaussian~\cite{lightgaussian} performs post-training pruning and subsequent fine-tuning to compress the Gaussian set while maintaining reconstruction quality. Speedy-Splat~\cite{speedysplat} optimizes the rendering pipeline 
for precise Gaussian localization and integrates a training-time pruning 
technique, jointly improving rendering speed, model compactness, and 
training efficiency.
Overall, these methods primarily optimize \emph{how many} Gaussians participate in rendering and learning, often improving model compactness and rendering efficiency; the impact on end-to-end training time depends on the specific pipeline and protocol.

\paragraph{Gaussian growth control.}
Another line of work targets the \emph{growth dynamics} of densification by regulating the Gaussian count under a resource budget, rather than removing Gaussians after the fact.
Taming 3DGS~\cite{mallick2024taming3dgs} studies training under compute/memory budgets and proposes mechanisms that control densification and optimization under constrained resources.
These methods operate at the representation-growth level and primarily optimize \emph{how large} the Gaussian set becomes (and thus the per-iteration workload), with some works explicitly reporting training-time speedups.

\paragraph{Our position: backward gating from the two-phase training structure.}
Existing efficiency improvements largely optimize the \emph{primitive dimension} of 3DGS training by either reducing the Gaussian set or controlling its growth during densification.
In contrast, we target a largely overlooked axis: \emph{whether} to execute the backward pass at each iteration.
SkipGS exploits per-view convergence heterogeneity in post-densification training and gates redundant backward passes based on per-view loss deviation, while enforcing stability budgets.

\section{Preliminaries: 3D Gaussian Splatting}
\label{sec:prelim}

3D Gaussian Splatting (3DGS)~\cite{kerbl3Dgaussians} represents a scene
as a set of $N$ learnable anisotropic 3D Gaussians and renders novel views
through a fully differentiable, tile-based rasterizer.
Each Gaussian $G_i$ is defined by a center
$\boldsymbol{\mu}_i\!\in\!\mathbb{R}^{3}$, a 3D covariance matrix
$\boldsymbol{\Sigma}_i\!\in\!\mathbb{R}^{3\times 3}$, an opacity
$\alpha_i\!\in\![0,1]$, and a set of spherical-harmonic (SH) coefficients
$\mathbf{f}_i$ encoding view-dependent color.
The Gaussian density at a point $\mathbf{x}$ is
\begin{equation}
  G_i(\mathbf{x})
  = \exp\!\Bigl(
      -\tfrac{1}{2}\,
      (\mathbf{x}-\boldsymbol{\mu}_i)^{\!\top}
      \boldsymbol{\Sigma}_i^{-1}\,
      (\mathbf{x}-\boldsymbol{\mu}_i)
    \Bigr).
  \label{eq:gaussian}
\end{equation}
To guarantee positive semi-definiteness during gradient-based optimization,
$\boldsymbol{\Sigma}_i$ is factorized as
\begin{equation}
  \boldsymbol{\Sigma}_i
  = \mathbf{R}_i\,\mathbf{S}_i\,\mathbf{S}_i^{\!\top}\mathbf{R}_i^{\!\top},
  \label{eq:cov}
\end{equation}
where $\mathbf{R}_i$ is a rotation matrix stored as a unit quaternion
$\mathbf{q}_i$ and
$\mathbf{S}_i=\mathrm{diag}(s_{i,1}, \allowbreak s_{i,2}, \allowbreak s_{i,3})$
is a diagonal scaling matrix whose entries represent the per-axis standard deviations of the Gaussian.

Given a camera with viewing transform $\mathbf{W}$ and the Jacobian
$\mathbf{J}$ of the local affine approximation to the projective
mapping~\cite{zwicker2001ewa}, the 2D screen-space covariance is
\begin{equation}
  \boldsymbol{\Sigma}_i^{\,2\mathrm{D}}
  = \mathbf{J}\,\mathbf{W}\,\boldsymbol{\Sigma}_i\,
    \mathbf{W}^{\!\top}\mathbf{J}^{\!\top}.
  \label{eq:proj}
\end{equation}
Pixel colors are composited by front-to-back alpha blending over the
depth-sorted Gaussians overlapping a given pixel $\mathbf{p}$:
\begin{equation}
  C(\mathbf{p})
  = \sum_{i\in\mathcal{N}}
      c_i\;\alpha_i'\;T_i,
  \qquad
  T_i = \prod_{j=1}^{i-1}\bigl(1-\alpha_j'\bigr),
  \label{eq:blend}
\end{equation}
where $c_i$ is the SH-evaluated color for the current view direction,
$T_i$ denotes the transmittance accumulated from all Gaussians in
front of $G_i$, and the effective per-pixel opacity is
\begin{equation}
  \alpha_i'
  = \alpha_i\,\exp\!\Bigl(
      -\tfrac{1}{2}\,
      \Delta\mathbf{p}_i^{\!\top}\,
      \bigl(\boldsymbol{\Sigma}_i^{\,2\mathrm{D}}\bigr)^{-1}
      \Delta\mathbf{p}_i
    \Bigr),
  \quad
  \Delta\mathbf{p}_i = \mathbf{p} - \boldsymbol{\mu}_i^{\,2\mathrm{D}}.
  \label{eq:alpha}
\end{equation}

The Gaussian parameters are initialized from a Structure-from-Motion~\cite{schoenberger2016sfm}
point cloud and optimized with
Adam~\cite{kingma2015adam} under a photometric loss
\begin{equation}
  \mathcal{L}
  = (1-\lambda)\,\mathcal{L}_1
  + \lambda\,\bigl(1-\mathrm{SSIM}\bigr),
  \label{eq:loss}
\end{equation}
where $\lambda=0.2$ and SSIM is the structural similarity
index~\cite{wang2004ssim}.
During training, an adaptive density control scheme periodically
\emph{clones} under-reconstructed Gaussians (high view-space positional
gradient, small spatial extent), \emph{splits} over-reconstructed ones
(high gradient, large extent), and \emph{prunes} near-transparent
Gaussians ($\alpha_i<\epsilon_\alpha$, where $\epsilon_\alpha$ is a small opacity threshold).

Crucially, 3DGS~\cite{kerbl3Dgaussians} training follows a two-phase training process induced by densification:
\begin{itemize}
  \item \textbf{Densification phase} (iterations $0$--$T_d$, $T_d{=}15$k by default):
        densification is active and the Gaussian count grows rapidly.
  \item \textbf{Post-densification phase} (iterations $T_d$--$T$, $T{=}30$k by default):
        densification is disabled; the Gaussian set is \emph{fixed} (typically $N=1$M--$5$M) and parameters are refined.
\end{itemize}

\section{Proposed Method}
\label{sec:method}

\subsection{Overview: Constrained Backward Gating}
\label{sec:method-overview}
Within this section, $t$ denotes the iteration index relative to the start of the post-densification phase (i.e., $t=1$ at the first iteration after $T_d$).
We focus on the post-densification phase.
In this phase, per-view losses often stabilize at different rates: a subset
of views continues to provide useful gradients while many others are
repeatedly revisited with little progress.
Yet standard 3DGS~\cite{kerbl3Dgaussians} still performs a full backward pass whenever a view is
sampled.
Since backward dominates the per-iteration cost, this update
redundancy in the late-phase leads to substantial wasted computation.
This observation motivates our backward gating mechanism, which decides for each sampled view whether to skip or execute the backward pass based on its recent loss deviation.

\begin{figure*}[!t]
  \centering
  \includegraphics[width=\linewidth]{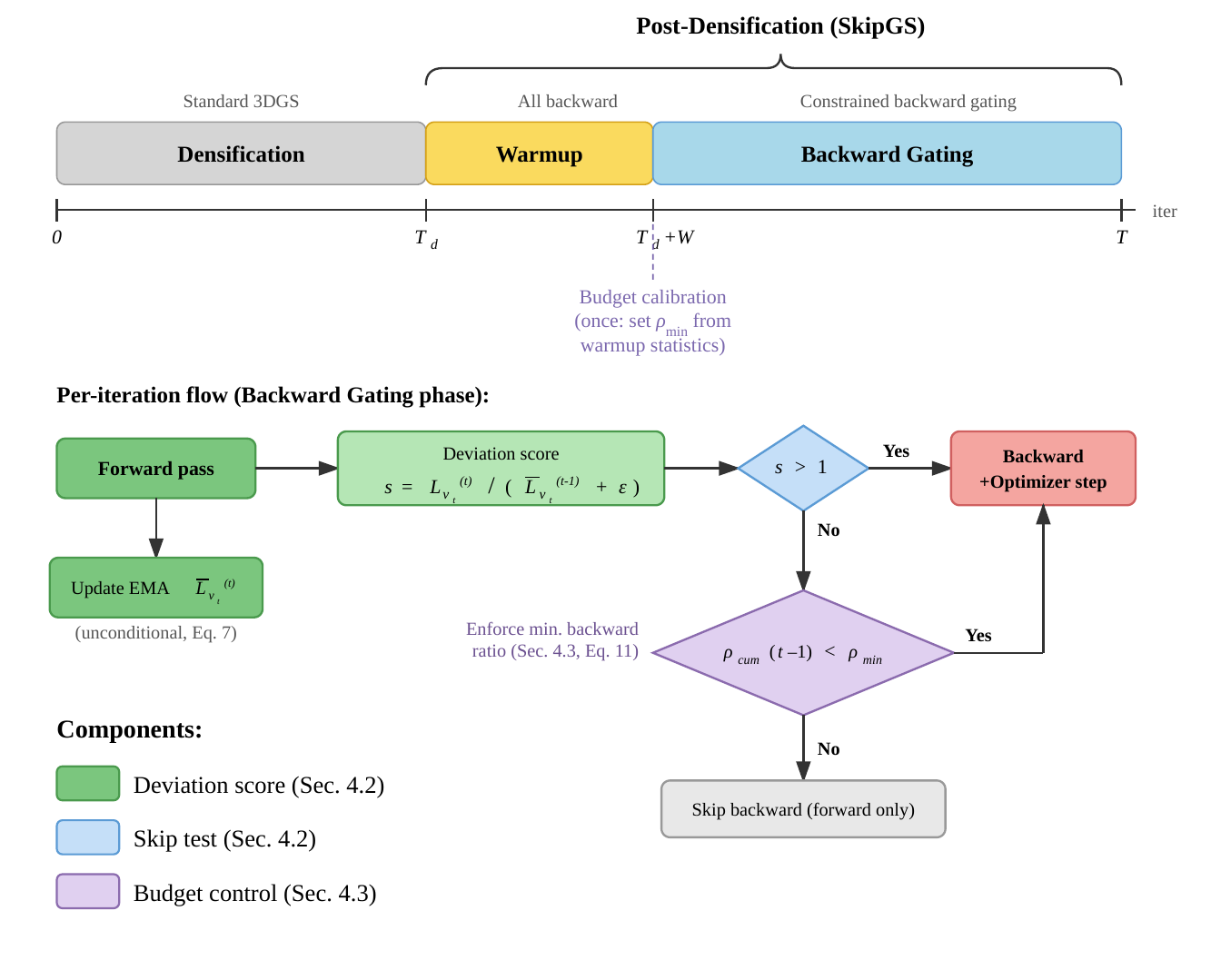}
  \caption{\textbf{Overview of SkipGS.}
  \emph{Top:} Training timeline. SkipGS activates after densification stops at $T_d$: a warmup window ($W$ iterations) initializes a per-view exponential moving average (EMA) baseline and calibrates the minimum backward budget $\rho_{\min}$, after which constrained backward gating begins.
  \emph{Bottom:} Per-iteration decision flow during the backward gating phase.
  At each iteration, the forward pass is always executed to compute the loss and update the per-view EMA $\bar{\mathcal{L}}_{v_t}^{(t)}$ (unconditional).
  A \textbf{deviation score} $s = \mathcal{L}_{v_t}^{(t)} / (\bar{\mathcal{L}}_{v_t}^{(t-1)} + \epsilon)$ measures whether the current loss exceeds the recent per-view baseline (Sec.~\ref{sec:convergence-signal}).
  If $s > 1$, the backward pass is executed; otherwise, SkipGS proposes to skip.
  Before skipping, the \textbf{budget controller} checks whether the cumulative backward ratio $\rho_{\mathrm{cum}}$ has fallen below the auto-calibrated minimum $\rho_{\min}$, and forces backward execution if so (Sec.~\ref{sec:budget}).
  Only when both checks allow skipping is the backward pass omitted.}
  \label{fig:method}
\end{figure*}

SkipGS targets \textbf{post-densification} training, where the Gaussian set is fixed and optimization mainly fine-tunes appearance and geometry.
Although many training views have already converged in this regime, standard 3DGS~\cite{kerbl3Dgaussians} still performs a full backward pass on every sampled view, even when the view's loss is stable relative to its recent per-view baseline.
We formulate post-densification acceleration as a \textbf{constrained backward gating} problem.
At iteration $t$, a view $v_t$ is sampled and we always perform the forward pass to compute
$\mathcal{L}_{v_t}^{(t)}$ and update per-view statistics.
We then decide whether to execute the backward pass so as to (i) focus compute on views with remaining optimization potential while (ii) maintaining stable convergence by enforcing a minimum update budget.

As illustrated in Fig.~\ref{fig:method}, SkipGS implements this gating mechanism with two ingredients:
\begin{enumerate}
  \item \textbf{Per-view deviation score and skip test} (Sec.~\ref{sec:convergence-signal}): For each sampled view, we compute a scale-invariant deviation score that compares its current loss against its recent EMA baseline. Backward is executed only when the score exceeds $1$, indicating the view's loss has risen above its baseline and is likely to benefit from gradient computation.
  \item \textbf{Backward budget calibration and control} (Sec.~\ref{sec:budget}): To prevent gradient starvation from over-aggressive skipping, we enforce a minimum backward ratio $\rho_{\min}$, auto-calibrated from warmup statistics. When the cumulative backward ratio $\rho_{\mathrm{cum}}$ falls below $\rho_{\min}$, the skip decision is overridden and backward is forced.
\end{enumerate}

Algorithm~\ref{alg:view-sched} summarizes the overall procedure.
Last but not least, SkipGS only changes \emph{when} to backpropagate; it does not modify the
renderer, the Gaussian representation, or the loss. It is therefore orthogonal to existing accelerations such as Gaussian compaction and methods that regulate Gaussian count growth during densification, and can be combined with them for additive speedups while maintaining rendering quality.

\begin{algorithm}[t]
  \caption{SkipGS: Post-Densification Backward Gating}
  \label{alg:view-sched}
  \footnotesize
  \begin{algorithmic}[1]
  \Require Total iterations $T$, densification stop iteration $T_d$, warmup length $W$, EMA decay $\beta$, stability constant $\epsilon$, budget floor $\rho_{\mathrm{lo}}$
  \State $T_{\text{post}} \leftarrow T - T_d$ 
  \State Initialize per-view EMA $\bar{\mathcal{L}}[\cdot] \leftarrow \varnothing$;\quad
         $\rho_{\min} \leftarrow \rho_{\mathrm{lo}}$;\quad
         backward count $b \leftarrow 0$
  \For{$t = 1$ \textbf{to} $T_{\text{post}}$}
    \State Sample view $v_t$; compute forward loss $\mathcal{L}_{v_t}^{(t)}$
    \State \textbf{Deviation score:} $s \leftarrow \mathcal{L}_{v_t}^{(t)} / (\bar{\mathcal{L}}_{v_t}^{(t-1)}+\epsilon)$ if $\bar{\mathcal{L}}_{v_t}^{(t-1)}$ exists, else $s \leftarrow +\infty$
    \State \textbf{Skip test:} $g \leftarrow \mathbf{1}[s > 1]$
    \State $\rho_{\mathrm{cum}}(t{-}1) \leftarrow b\,/\,\max(t{-}1,\,1)$ \Comment{cumulative backward ratio before iteration $t$}
    \If{$t \le W$} \Comment{warmup: always backward, collect stats}
      \State Record $g$ for budget calibration; force $g \leftarrow 1$
    \Else
      \State Calibrate $\rho_{\min}$ from warmup stats (once, at $t{=}W{+}1$; Eq.~\ref{eq:rhomincalib})
      \State If $\rho_{\mathrm{cum}}(t{-}1) < \rho_{\min}$, force $g \leftarrow 1$ \Comment{budget override}
    \EndIf
    \If{$g = 1$}
      \State Backward; optimizer step; $b \leftarrow b + 1$
    \EndIf
    \State Update $\bar{\mathcal{L}}_{v_t}^{(t)} \leftarrow \beta\,\bar{\mathcal{L}}_{v_t}^{(t-1)} + (1{-}\beta)\,\mathcal{L}_{v_t}^{(t)}$ \Comment{unconditional (Eq.~\ref{eq:view-ema})}
  \EndFor
  \end{algorithmic}
\end{algorithm}

\subsection{Per-View Deviation Score}
\label{sec:convergence-signal}

After densification is disabled, different views converge at different rates due to varying
visibility, occlusion patterns, and view-dependent appearance. A single global loss trend is
therefore insufficient to identify which sampled views still provide useful optimization signal.
We construct a lightweight per-view deviation score from recent loss statistics.

For each training view, we maintain an exponential moving average (EMA) of its observed loss. When view $v_t$ is sampled at iteration $t$,
\begin{equation}
  \bar{\mathcal{L}}_{v_t}^{(t)} =
  \beta\, \bar{\mathcal{L}}_{v_t}^{(t-1)} + (1-\beta)\,\mathcal{L}_{v_t}^{(t)},
  \label{eq:view-ema}
\end{equation}
with $\beta\in(0,1)$; unsampled views keep their previous EMA.
We then measure the normalized deviation of $v_t$ from this baseline by a scale-invariant ratio:
\begin{equation}
s_{v_t}^{(t)} = \frac{\mathcal{L}_{v_t}^{(t)}}{\bar{\mathcal{L}}_{v_t}^{(t-1)}+\epsilon},
  \label{eq:deviation}
\end{equation}
where $\epsilon=10^{-8}$ is a small constant for numerical stability.
In implementation, we update the per-view EMA after each forward observation and deviation score calculation, regardless of whether the backward pass is executed.
Thus, $\mathcal{L}_{v_t}^{(t)}$ always denotes the observed forward loss, ensuring the deviation score tracks per-view loss trends
independent of gradient updates.

We refer to $s_{v_t}^{(t)}$ as the \emph{deviation score}. Values significantly above $1$ indicate a loss spike
relative to the recent baseline, suggesting that the view remains under-optimized (or has regressed)
and is likely to benefit from backward computation. Values below $1$ indicate the view currently performs
better than its recent baseline; in this case, backward passes typically yield diminishing
returns, and we prioritize compute for high-deviation views.

Given the deviation score, we decide whether to execute backward via a skip test:
\begin{equation}
  g_t(v_t)=\mathbf{1}\!\left[s_{v_t}^{(t)} > 1\right],
  \label{eq:gate}
\end{equation}
where $g_t(v_t)=1$ triggers a backward pass and $g_t(v_t)=0$ skips it.
We backpropagate only when the view's current loss exceeds its recent EMA baseline, i.e., when the deviation score is above $1$.

To allow per-view EMAs to stabilize after densification ends, we apply a warmup window of $W$ iterations
during which all backward passes are unconditionally executed.
During warmup, we still evaluate the skip test for budget calibration (Sec.~\ref{sec:budget}) and collect statistics
only when the sampled view already has an EMA history.

\subsection{Backward Budget Calibration and Control}
\label{sec:budget}

The view-adaptive backward gating in Eq.~\eqref{eq:gate} proposes skipping on low-utility views, but unconstrained skipping can starve optimization of gradient signal.
We therefore enforce a \textbf{minimum backward budget} on average: when the cumulative backward ratio since the start of the post-densification phase falls below a target minimum, we override the gating decision and force backward execution.

Let $g_t\in\{0,1\}$ indicate whether post-densification iteration $t$ executes backward.
We maintain the cumulative backward ratio
\begin{equation}
  \rho_{\mathrm{cum}}(t) = \frac{1}{t}\sum_{i=1}^{t} g_i.
  \label{eq:rhocum}
\end{equation}
Since $g_t$ is decided online, we apply the budget-floor check using the ratio \emph{before} making the current decision:
\begin{equation}
  \text{if }\rho_{\mathrm{cum}}(t{-}1) < \rho_{\min}\text{ then force } g_t \leftarrow 1.
  \label{eq:rhocum_check}
\end{equation}
Because warmup iterations execute backward ($g_t{=}1$), including warmup in $\rho_{\mathrm{cum}}$ yields a smooth budget signal that is less sensitive to short-term fluctuations.

The appropriate minimum budget depends on scene difficulty and densification quality.
To avoid per-scene tuning, we calibrate $\rho_{\min}$ from warmup statistics.
During warmup (backward always executed), we still evaluate the gating criterion in Eq.~\eqref{eq:gate} and record the fraction of iterations that would trigger backward:
\begin{equation}
  \hat{\rho}_{W}=\frac{1}{|\mathcal{T}_{W}|}\sum_{i\in\mathcal{T}_{W}} g_i(v_i),
  \label{eq:rhohat}
\end{equation}
where $\mathcal{T}_{W}$ denotes warmup iterations whose sampled view already has an EMA history.
Equivalently, $1 - \hat{\rho}_{W}$ gives the fraction of warmup iterations where the gating would have proposed skipping ($s \le 1$).

We then set the minimum backward ratio by a lower-bounded linear interpolation:
\begin{equation}
  \rho_{\min}= \rho_{\mathrm{lo}} + (1-\rho_{\mathrm{lo}})\,\hat{\rho}_{W},
  \label{eq:rhomincalib}
\end{equation}
where $\rho_{\mathrm{lo}}=0.5$ in all experiments.
Intuitively, if warmup indicates many views are already easy (small $\hat{\rho}_{W}$), we allow a lower minimum budget; if warmup suggests most views still require backward ($\hat{\rho}_{W}\approx 1$), we enforce a budget close to full backward.

\section{Experiments}
\label{sec:experiments}

We evaluate \textit{SkipGS} on standard 3DGS~\cite{kerbl3Dgaussians} benchmarks to assess its ability to
reduce wall-clock training time in the post-densification phase while preserving reconstruction quality.
We report results on the vanilla 3DGS pipeline and on representative acceleration baselines that either
(i) reduce the number of Gaussians (pruning/compaction) or (ii) control Gaussian growth during densification.
For each baseline, we additionally report the performance of \textit{baseline + SkipGS (ours)} to quantify
the additive benefit and verify composability.

\subsection{Experimental Setup}
\label{sec:exp-setup}

\paragraph{Datasets.}
Following the standard 3DGS~\cite{kerbl3Dgaussians} protocol, we conduct experiments on three real-world datasets:
Mip-NeRF 360~\cite{barron2022mipnerf360}, Deep Blending~\cite{hedman2018deepblending}, and
Tanks and Temples~\cite{Knapitsch2017tanks}.
Unless otherwise specified, we train each scene for $30$K iterations with densification enabled until $T_d=15$K,
followed by post-densification fine-tuning.

\paragraph{Baselines and evaluation protocol.}
We compare against vanilla 3DGS~\cite{kerbl3Dgaussians} and representative efficiency-oriented baselines from two families:
(i) \emph{Gaussian compaction} methods that reduce the number of Gaussians (e.g., FastGS~\cite{feng2024fastgs},
GaussianSpa~\cite{gaussianspa}, LightGaussian~\cite{lightgaussian}, Speedy-Splat~\cite{speedysplat}); and
(ii) \emph{Gaussian growth control} methods that regulate densification under constraints (e.g., Taming 3DGS~\cite{mallick2024taming3dgs}).
For each baseline, we evaluate both the original method and the combined setting \textit{baseline + SkipGS},
where our method is enabled only in the post-densification phase and all other components of the baseline remain unchanged.

\paragraph{Metrics.}
We report standard novel-view synthesis metrics, including PSNR, SSIM~\cite{wang2004ssim}, and LPIPS~\cite{Zhang_2018_CVPR}.
Training efficiency is evaluated by the end-to-end wall-clock training time ($T_{\mathrm{total}}$) as well as the post-densification refinement time ($T_{\mathrm{post}}$), both measured in seconds.

\paragraph{Implementation details.}
All experiments are run on a single NVIDIA RTX Ada 5000 (32\,GB) GPU.
Unless a baseline requires otherwise, we use the standard 3DGS~\cite{kerbl3Dgaussians} optimizer (Adam) and loss as in~\cite{kerbl3Dgaussians}.
SkipGS is enabled only after densification stops (post-densification phase): we always execute the forward pass,
maintain per-view loss EMAs, and decide whether to execute backward via the view-adaptive backward gating (Eq.~\eqref{eq:gate}).
We use the same hyperparameters across all scenes and datasets in our evaluation:
warmup length $W{=}500$, EMA decay $\beta{=}0.95$, stability constant $\epsilon{=}10^{-8}$,
and $\rho_{\min}$ auto-calibrated at the end of warmup from the observed natural convergence rate of each scene.
When backward is executed, we apply the standard optimizer update.
We report end-to-end wall-clock speedups for the full 30K-iteration training unless stated otherwise.

\subsection{Main Results: Training Acceleration in Post-Densification}
\label{sec:exp-main}

\begin{table}[!t]
  \centering
    \caption{\textbf{Mip-NeRF~360~\cite{barron2022mipnerf360} (avg over scenes):} comparisons with 3DGS baselines and \textit{baseline + SkipGS}.
    $T_{\text{total}}$ is the end-to-end wall-clock training time (s) under each method's protocol.
    $T_{\text{post}}$ denotes the post-densification refinement time (s).
    LightGaussian performs prune+finetune after completing a full 3DGS training run; therefore we report only this additional post-training, so $T_{\text{total}}=T_{\text{post}}$ for LightGaussian rows. 
    Note that SkipGS does not modify the renderer or the Gaussian set; hence the number of Gaussians as well as the rendering-time workload are identical to the corresponding baseline, and our speedups come solely from reducing post-densification backpropagation.}

\label{tab:main_mip}
  \scriptsize
  \setlength{\tabcolsep}{4pt}
  \renewcommand{\arraystretch}{1.12}
  \begin{tabular}{lccccc}
    \toprule
    Method & PSNR$\uparrow$ & SSIM$\uparrow$ & LPIPS$\downarrow$ & $T_{\text{total}}\downarrow$ & $T_{\text{post}}\downarrow$ \\
    \midrule
    Vanilla 3DGS
      & 27.52 & 0.816 & 0.215 & 1705.7 & 939.6 \\
    + SkipGS (ours)
      & 27.52 & 0.816 & 0.217 & 1311.1 & 545.0 \\
    $\Delta$ (Ours--Vanilla)
      & +0.00 & 0.000 & +0.002 & -394.6 (-23.1\%) & -394.6 (-42.0\%) \\
    \midrule
    FastGS~\cite{feng2024fastgs}
      & 27.56 & 0.798 & 0.261 & 181.9 & 87.2 \\
    + SkipGS (ours)
      & 27.51 & 0.797 & 0.262 & \textbf{164.5} & \textbf{69.8} \\
    $\Delta$ (Ours--FastGS)
      & -0.05 & -0.001 & +0.001 & -17.4 (-9.6\%) & -17.4 (-20.0\%) \\
    \midrule
    Taming 3DGS~\cite{mallick2024taming3dgs}
      & 27.94 & 0.822 & 0.207 & 1339.0 & 757.0 \\
    + SkipGS (ours)
      & 27.92 & 0.822 & 0.209 & 974.0 & 392.0 \\
    $\Delta$ (Ours--Taming 3DGS)
      & -0.02 & 0.000 & +0.002 & -365.0 (-27.3\%) & -365.0 (-48.2\%) \\
    \midrule
    GaussianSpa~\cite{gaussianspa}
      & 27.61 & 0.826 & 0.213 & 2640.9 & 1485.0 \\
    + SkipGS (ours)
      & 27.60 & 0.825 & 0.215 & 2490.9 & 1335.0 \\
    $\Delta$ (Ours--GaussianSpa)
      & -0.01 & -0.001 & +0.002 & -150.0 (-5.7\%) & -150.0 (-10.1\%) \\
    \midrule
    LightGaussian~\cite{lightgaussian}
      & 27.49 & 0.810 & 0.230 & 240.0 & 240.0 \\
    + SkipGS (ours)
      & 27.46 & 0.809 & 0.231 & 204.8 & 204.8 \\
    $\Delta$ (Ours--LightGaussian)
      & -0.03 & -0.001 & +0.001 & -35.2 (-14.7\%) & -35.2 (-14.7\%) \\
    \midrule
    Speedy-Splat~\cite{speedysplat}
      & 27.11 & 0.799 & 0.263 & 1099.8 & 492.0 \\
    + SkipGS (ours)
      & 27.08 & 0.799 & 0.264 & 1030.8 & 423.0 \\
    $\Delta$ (Ours--Speedy-Splat)
      & -0.03 & 0.000 & +0.001 & -69.0 (-6.3\%) & -69.0 (-14.0\%) \\
    \bottomrule
  \end{tabular}
\end{table}

\begin{table}[!t]
  \centering
    \caption{\textbf{Deep Blending~\cite{hedman2018deepblending} (avg over scenes):} same protocol and definitions as Table~\ref{tab:main_mip}. For this dataset, we compute $T_{\text{total}}$ for our method as the sum of the baseline method’s densification-phase wall-clock time and our $T_{\text{post}}$.}

\label{tab:main_db}
  \scriptsize
  \setlength{\tabcolsep}{4pt}
  \renewcommand{\arraystretch}{1.12}
  \begin{tabular}{lccccc}
    \toprule
    Method & PSNR$\uparrow$ & SSIM$\uparrow$ & LPIPS$\downarrow$ & $T_{\text{total}}\downarrow$ & $T_{\text{post}}\downarrow$ \\
    \midrule
    Vanilla 3DGS
      & 29.79 & 0.907 & 0.238 & 1713.0 & 965.5 \\
    + SkipGS (ours)
      & 29.88 & 0.909 & 0.237 & 1349.5 & 602.0 \\
    $\Delta$ (Ours--Vanilla)
      & +0.09 & +0.002 & -0.001 & -363.5 (-21.2\%) & -363.5 (-37.6\%) \\
    \midrule
    FastGS~\cite{feng2024fastgs}
      & 30.02 & 0.905 & 0.267 & 111.0 & 49.6 \\
    + SkipGS (ours)
      & 30.03 & 0.906 & 0.268 & \textbf{100.6} & \textbf{39.2} \\
    $\Delta$ (Ours--FastGS)
      & +0.01 & +0.001 & +0.001 & -10.4 (-9.4\%) & -10.4 (-21.0\%) \\
    \midrule
    Taming 3DGS~\cite{mallick2024taming3dgs}
      & 29.83 & 0.907 & 0.237 & 1110.0 & 620.0 \\
    + SkipGS (ours)
      & 29.95 & 0.909 & 0.237 & 804.0 & 314.0 \\
    $\Delta$ (Ours--Taming 3DGS)
      & +0.12 & +0.002 & 0.000 & -306.0 (-27.6\%) & -306.0 (-49.4\%) \\
    \midrule
    GaussianSpa~\cite{gaussianspa}
      & 30.19 & 0.913 & 0.238 & 2373.5 & 1309.5 \\
    + SkipGS (ours)
      & 30.27 & 0.916 & 0.236 & 2251.5 & 1187.5 \\
    $\Delta$ (Ours--GaussianSpa)
      & +0.08 & +0.003 & -0.002 & -122.0 (-5.1\%) & -122.0 (-9.3\%) \\
    \midrule
    LightGaussian~\cite{lightgaussian}
      & 29.92 & 0.905 & 0.251 & 240.0 & 240.0 \\
    + SkipGS (ours)
      & 29.92 & 0.906 & 0.252 & 204.5 & 204.5 \\
    $\Delta$ (Ours--LightGaussian)
      & 0.00 & +0.001 & +0.001 & -35.5 (-14.8\%) & -35.5 (-14.8\%) \\
    \midrule
    Speedy-Splat~\cite{speedysplat}
      & 29.60 & 0.905 & 0.260 & 977.5 & 441.5 \\
    + SkipGS (ours)
      & 29.59 & 0.905 & 0.261 & 916.0 & 380.0 \\
    $\Delta$ (Ours--Speedy-Splat)
      & -0.01 & 0.000 & +0.001 & -61.5 (-6.3\%) & -61.5 (-13.9\%) \\
    \bottomrule
  \end{tabular}
\end{table}

\begin{table}[!t]
  \centering
    \caption{\textbf{Tanks\&Temples~\cite{Knapitsch2017tanks} (avg over scenes):} same protocol and definitions as Table~\ref{tab:main_mip}. For this dataset, we compute $T_{\text{total}}$ for our method as the sum of the baseline method's densification-phase wall-clock time and our $T_{\text{post}}$.}
  \label{tab:main_tt}
  \scriptsize
  \setlength{\tabcolsep}{4pt}
  \renewcommand{\arraystretch}{1.12}
  \begin{tabular}{lccccc}
    \toprule
    Method & PSNR$\uparrow$ & SSIM$\uparrow$ & LPIPS$\downarrow$ & $T_{\text{total}}\downarrow$ & $T_{\text{post}}\downarrow$ \\
    \midrule
    Vanilla 3DGS
      & 23.83 & 0.853 & 0.169 & 993.0 & 545.5 \\
    + SkipGS (ours)
      & 23.78 & 0.853 & 0.173 & 792.0 & 344.5 \\
    $\Delta$ (Ours--Vanilla)
      & -0.05 & 0.000 & +0.004 & -201.0 (-20.2\%) & -201.0 (-36.8\%) \\
    \midrule
    FastGS~\cite{feng2024fastgs}
      & 24.25 & 0.843 & 0.208 & 106.0 & 46.1 \\
    + SkipGS (ours)
      & 24.17 & 0.842 & 0.210 & \textbf{98.3} & \textbf{38.4} \\
    $\Delta$ (Ours--FastGS)
      & -0.08 & -0.001 & +0.002 & -7.7 (-7.3\%) & -7.7 (-16.7\%) \\
    \midrule
    Taming 3DGS~\cite{mallick2024taming3dgs}
      & 24.43 & 0.859 & 0.164 & 774.0 & 424.0 \\
    + SkipGS (ours)
      & 24.34 & 0.858 & 0.169 & 596.0 & 246.0 \\
    $\Delta$ (Ours--Taming 3DGS)
      & -0.09 & -0.001 & +0.005 & -178.0 (-23.0\%) & -178.0 (-42.0\%) \\
    \midrule
    GaussianSpa~\cite{gaussianspa}
      & 23.71 & 0.854 & 0.168 & 1675.5 & 731.5 \\
    + SkipGS (ours)
      & 23.66 & 0.854 & 0.169 & 1600.5 & 656.5 \\
    $\Delta$ (Ours--GaussianSpa)
      & -0.05 & 0.000 & +0.001 & -75.0 (-4.5\%) & -75.0 (-10.3\%) \\
    \midrule
    LightGaussian~\cite{lightgaussian}
      & 23.89 & 0.847 & 0.187 & 160.0 & 160.0 \\
    + SkipGS (ours)
      & 23.85 & 0.846 & 0.189 & 140.5 & 140.5 \\
    $\Delta$ (Ours--LightGaussian)
      & -0.04 & -0.001 & +0.002 & -19.5 (-12.2\%) & -19.5 (-12.2\%) \\
    \midrule
    Speedy-Splat~\cite{speedysplat}
      & 23.66 & 0.830 & 0.222 & 600.5 & 267.5 \\
    + SkipGS (ours)
      & 23.60 & 0.829 & 0.223 & 568.2 & 235.2 \\
    $\Delta$ (Ours--Speedy-Splat)
      & -0.06 & -0.001 & +0.001 & -32.3 (-5.4\%) & -32.3 (-12.1\%) \\
    \bottomrule
  \end{tabular}
\end{table}

We first evaluate SkipGS on the vanilla 3DGS pipeline~\cite{kerbl3Dgaussians}.
Across all datasets, SkipGS reduces end-to-end wall-clock training time by skipping redundant backward passes in the post-densification phase,
while preserving reconstruction quality measured by PSNR/SSIM/LPIPS.
On Mip-NeRF~360 (Table~\ref{tab:main_mip}), compared to vanilla 3DGS, SkipGS reduces the end-to-end training time by 23.1\%,
driven by a 42.0\% reduction in post-densification time $T_{\text{post}}$.
Please note that we report $T_{\text{total}}=T_{\text{post}}$ for the LightGaussian rows because LightGaussian performs prune+finetune after completing a full 3DGS training run.
Notably, the final number of Gaussians is unchanged (2.739M on average), confirming that the gains come from backward gating rather than primitive reduction.

%
Tables~\ref{tab:main_db} and~\ref{tab:main_tt} report the same comparison on Deep Blending and Tanks\&Temples.
We observe consistent reductions in $T_{\text{post}}$, which translate to end-to-end speedups without changing the Gaussian set.

Notably, FastGS\,+\,SkipGS achieves the fastest post-densification training across all three benchmarks (e.g., $T_{\text{post}}{=}69.8$\,s on Mip-NeRF\,360), demonstrating that SkipGS can further accelerate the current fastest 3DGS training pipeline.

\begin{figure*}[t]
  \centering
  \setlength{\tabcolsep}{1.5pt}
  \renewcommand{\arraystretch}{0.8}
  \begin{tabular}{ccc}
    
    \includegraphics[width=0.32\linewidth]{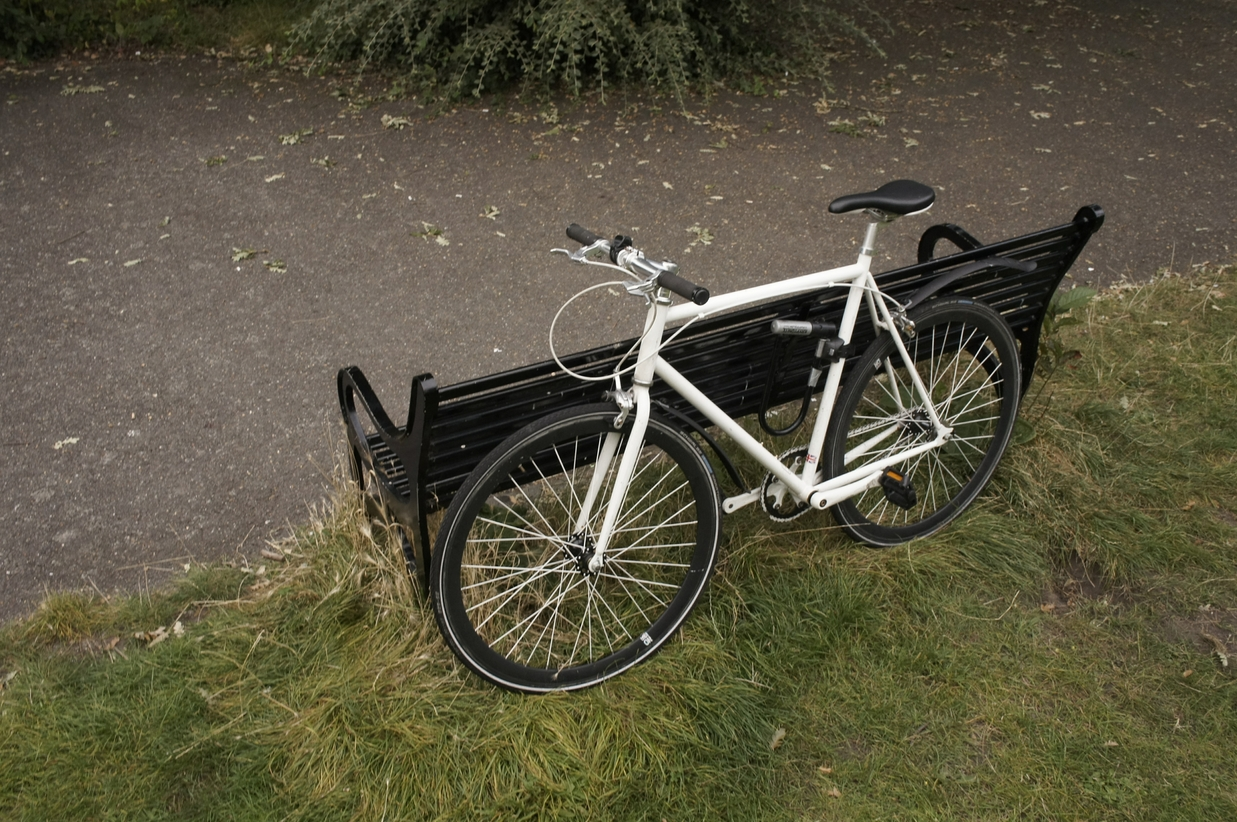} &
    \includegraphics[width=0.32\linewidth]{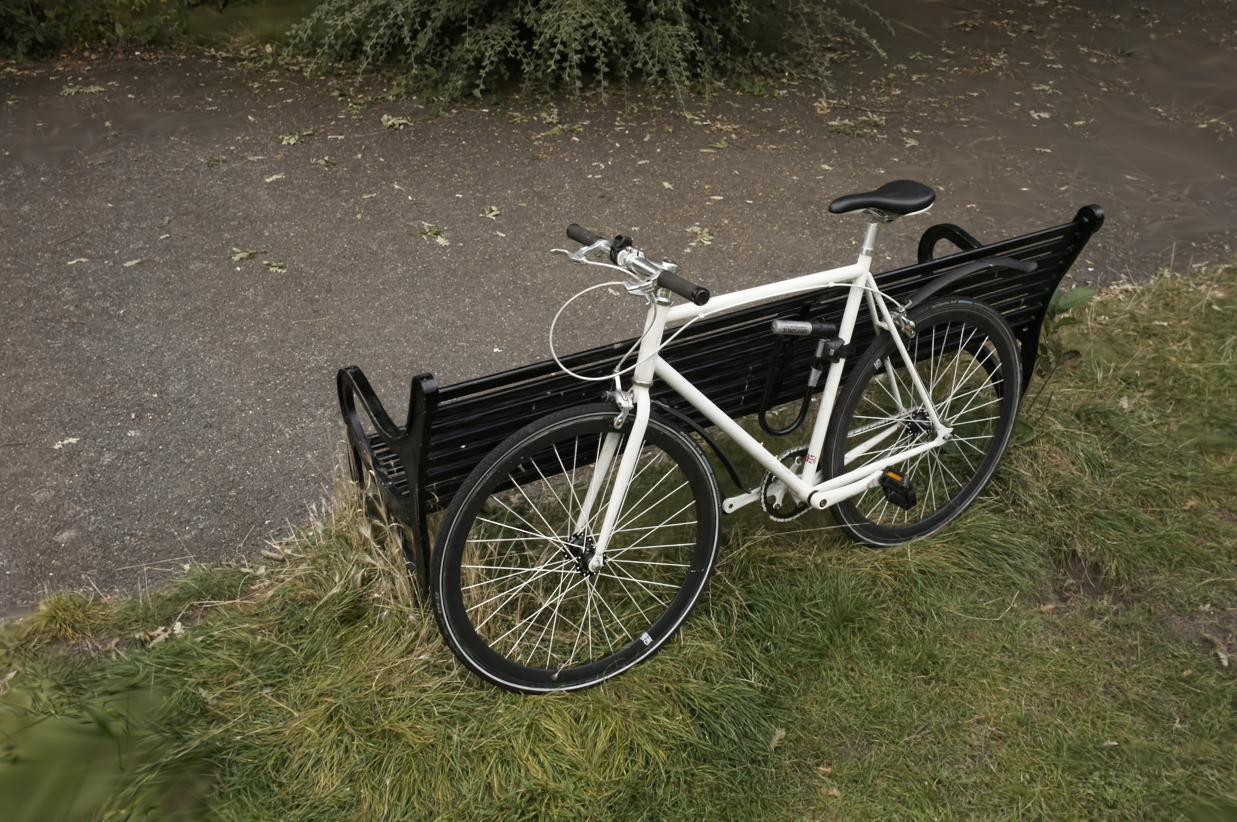} &
    \includegraphics[width=0.32\linewidth]{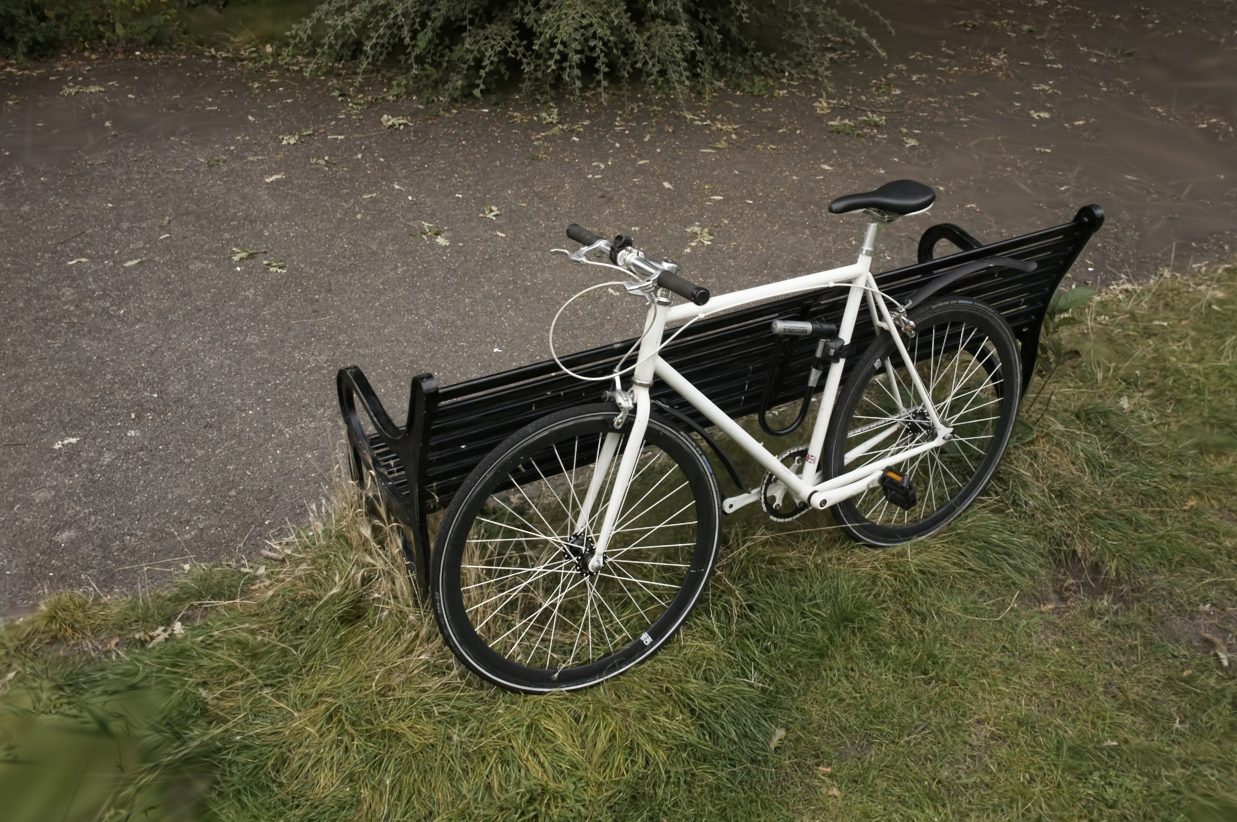} \\[-2pt]
    \multicolumn{3}{l}{\scriptsize\textit{bicycle} (Mip-NeRF\,360~\cite{barron2022mipnerf360}) --- Vanilla 3DGS, $T_{\text{post}}$: 939.6\,s $\rightarrow$ 545.0\,s (\textbf{--42.0\%})} \\[4pt]

    \includegraphics[width=0.32\linewidth]{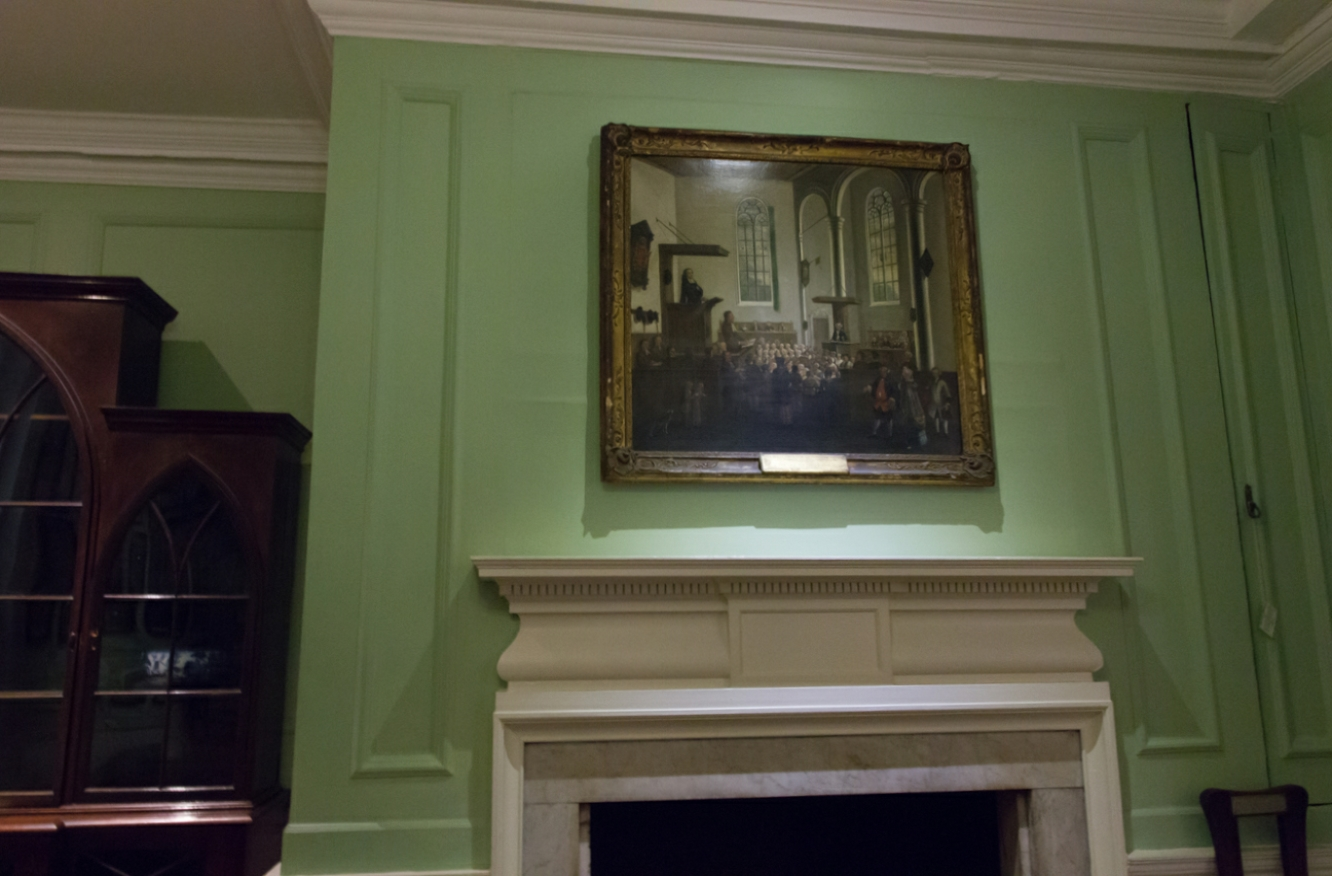} &
    \includegraphics[width=0.32\linewidth]{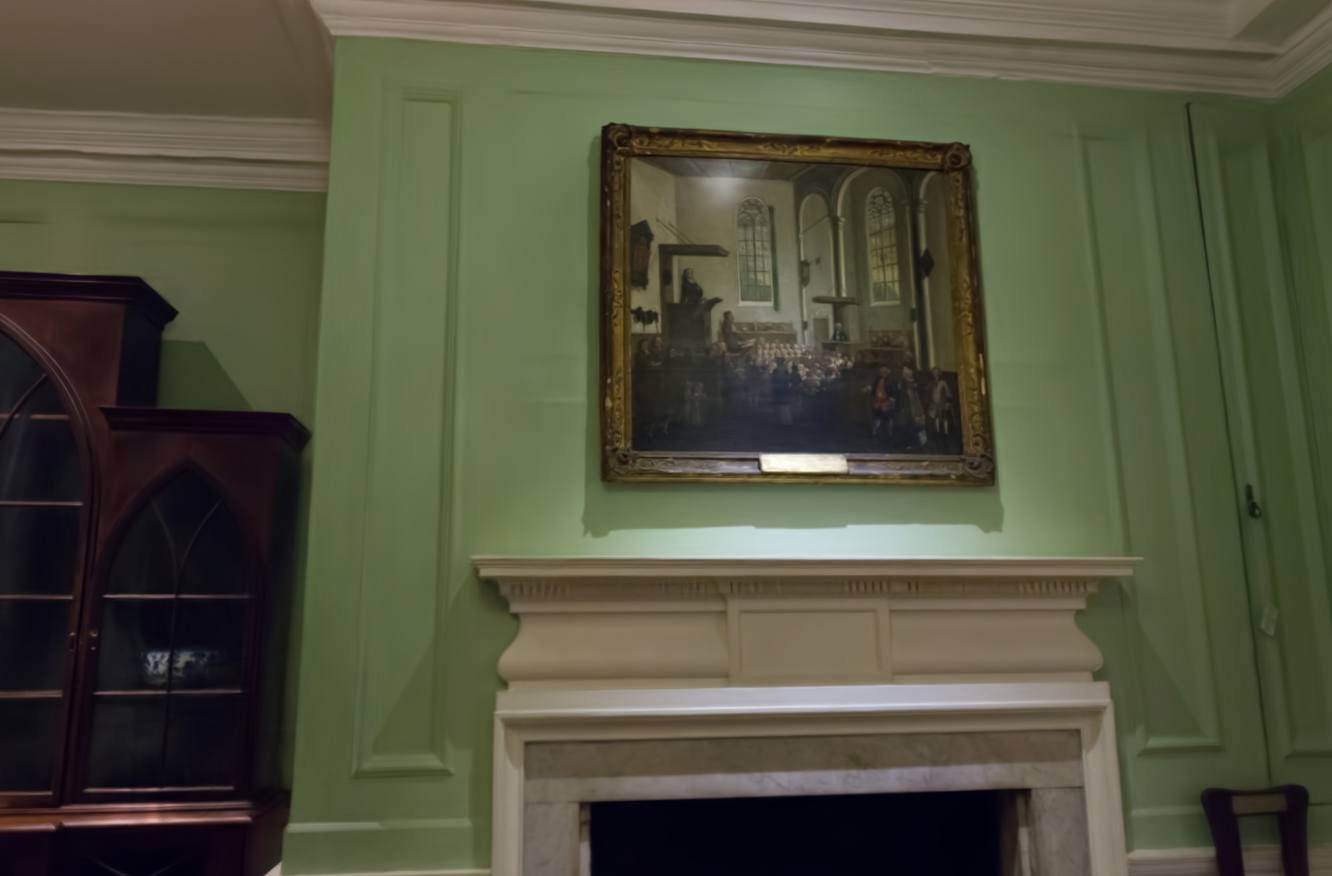} &
    \includegraphics[width=0.32\linewidth]{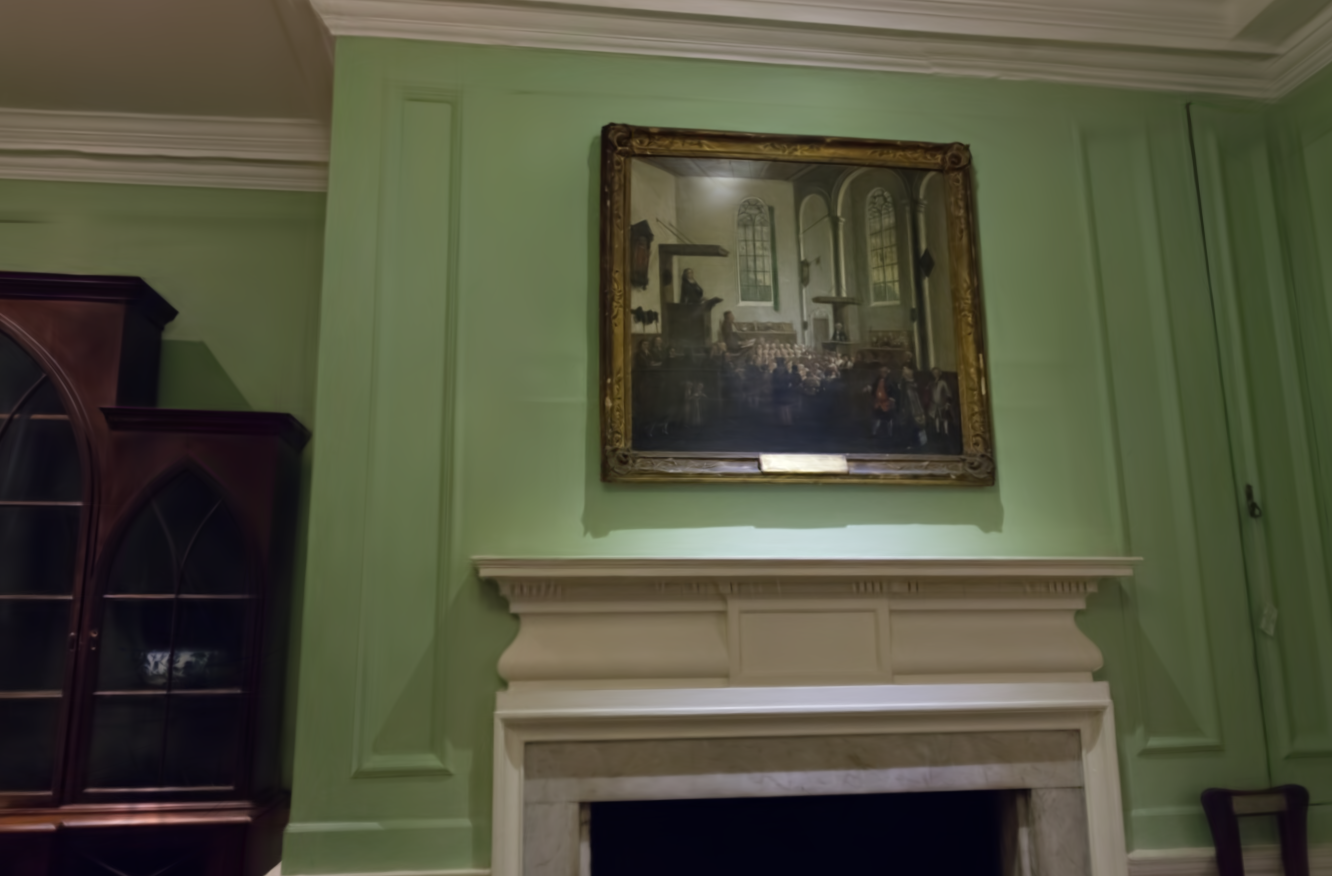} \\[-2pt]
    \multicolumn{3}{l}{\scriptsize\textit{drjohnson} (Deep Blending~\cite{hedman2018deepblending}) --- Taming 3DGS, $T_{\text{post}}$: 620\,s $\rightarrow$ 314\,s (\textbf{--49.4\%})} \\[4pt]

    \includegraphics[width=0.32\linewidth]{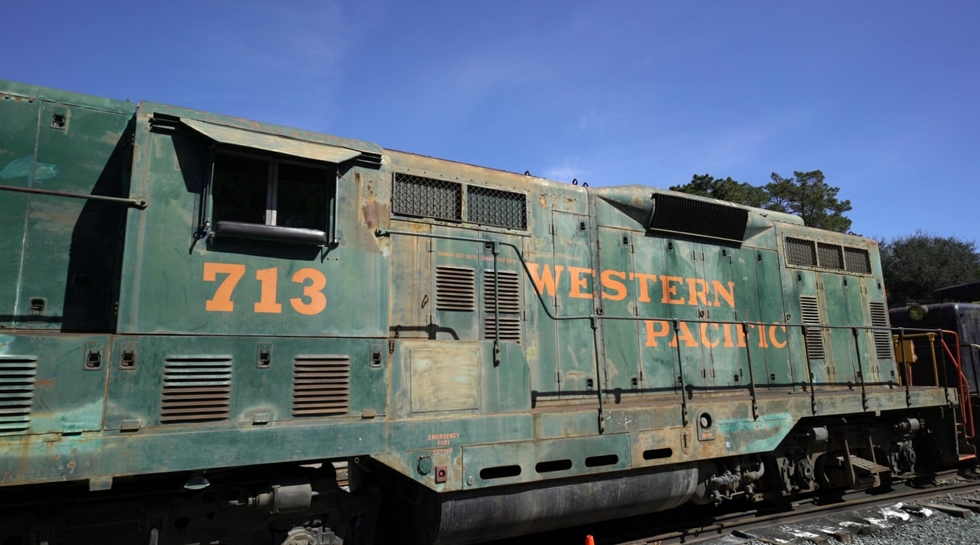} &
    \includegraphics[width=0.32\linewidth]{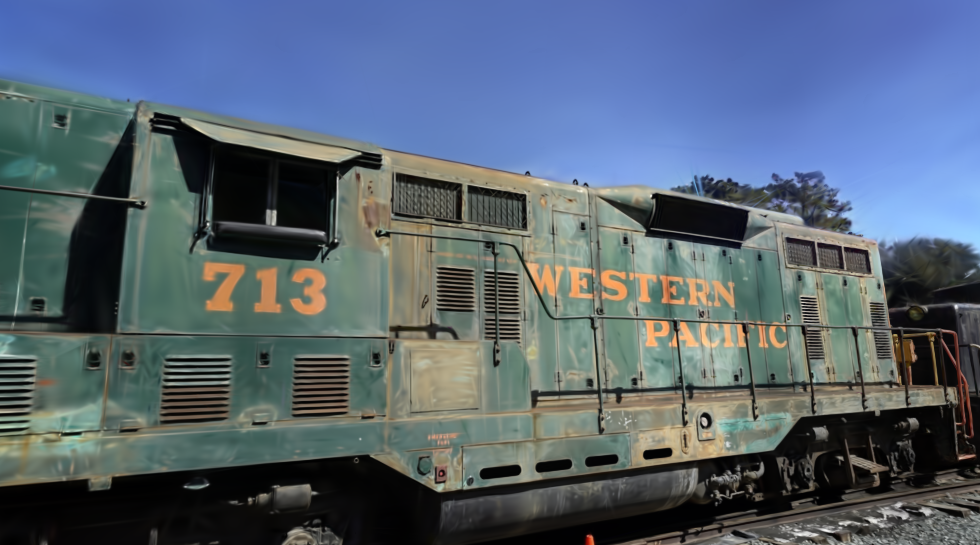} &
    \includegraphics[width=0.32\linewidth]{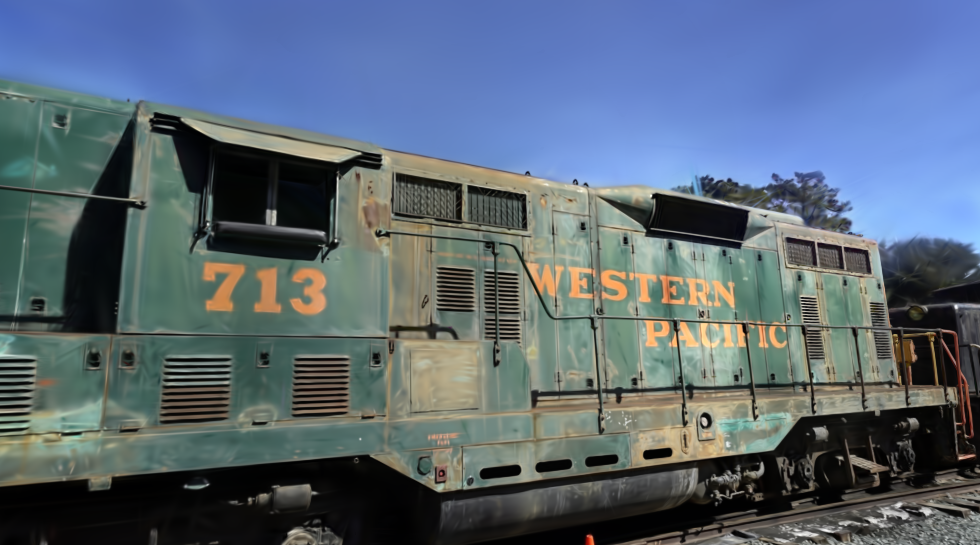} \\[-2pt]
    \multicolumn{3}{l}{\scriptsize\textit{train} (Tanks\&Temples~\cite{Knapitsch2017tanks}) --- Speedy-Splat, $T_{\text{post}}$: 267.5\,s $\rightarrow$ 235.2\,s (\textbf{--12.1\%})} \\[2pt]

    Ground Truth & Baseline & + SkipGS (Ours) \\
  \end{tabular}
  \caption{\textbf{Qualitative comparison across datasets and baselines.}
  Each row shows a different dataset and baseline method.
  Despite substantial reductions in post-densification time,
  SkipGS produces visually indistinguishable results from
  the corresponding full-training baseline across all settings.}
  \label{fig:qualitative}
\end{figure*}

Beyond vanilla 3DGS, we apply SkipGS as a post-densification plug-in on top of representative baselines that
(i) reduce or compact the Gaussian set (e.g., pruning/sparsification/compaction) or
(ii) regulate Gaussian growth during densification under resource constraints.
Tables~\ref{tab:main_mip}--\ref{tab:main_tt} report each baseline and its \textit{baseline + SkipGS} counterpart.
Across these baselines, SkipGS provides additional wall-clock savings by targeting an orthogonal axis:
it reduces \emph{how often} backward is executed in post-densification training, whereas prior methods mainly affect
\emph{how many} primitives are processed per iteration and/or memory/throughput constraints.
Importantly, SkipGS does not modify the renderer, representation, or loss, and thus can be enabled without re-tuning baseline-specific knobs;
the combined setting retains comparable reconstruction quality while improving training speed.

\paragraph{Why post-densification backward scheduling yields wall-clock gains.}
Fig.~\ref{fig:motivation}(a) shows that the backward pass accounts for the majority of per-iteration time throughout the post-densification phase.
Meanwhile, Fig.~\ref{fig:motivation}(b) shows that average per-Gaussian gradient norms flatten after $T_d$ while average Adam update norms remain stable due to momentum inertia, indicating that many post-densification backward passes produce weakly informative gradients.
SkipGS exploits this post-densification redundancy by always performing the forward pass to update per-view statistics,
but executing backward selectively under a minimum-budget constraint, thereby reducing the dominant post-densification cost in practice.

\paragraph{Qualitative results.}
Fig.~\ref{fig:qualitative} shows rendered views across three datasets and representative baselines.
Despite substantial post-densification time reductions, SkipGS produces visually indistinguishable results from the corresponding full-training baseline, consistent with Tables~\ref{tab:main_mip}--\ref{tab:main_tt}.

\subsection{Ablation Study}
\label{sec:ablation}

We ablate the backward budget control mechanism (Sec.~\ref{sec:budget}) on Mip-NeRF\,360~\cite{barron2022mipnerf360} using two representative baselines: GaussianSpa~\cite{gaussianspa} and Taming 3DGS~\cite{mallick2024taming3dgs}. Table~\ref{tab:ablation} compares three configurations for each: the baseline without SkipGS, the full SkipGS pipeline, and SkipGS with budget control removed.

\begin{table}[t]
  \centering
  \caption{\textbf{Ablation on Mip-NeRF\,360~\cite{barron2022mipnerf360} (avg over scenes).} Removing budget control leads to over-aggressive skipping: $T_{\text{post}}$ drops further but quality degrades substantially. The budget mechanism trades a modest time increase for quality preservation.}
  \label{tab:ablation}
  \scriptsize
  \setlength{\tabcolsep}{4pt}
  \renewcommand{\arraystretch}{1.15}
  \begin{tabular}{lcccc}
    \toprule
    Method & PSNR$\uparrow$ & SSIM$\uparrow$ & LPIPS$\downarrow$ & $T_{\text{post}}$(s)$\downarrow$ \\
    \midrule
    GaussianSpa (baseline) & 27.61 & 0.826 & 0.213 & 1485.0 \\
    + SkipGS (full) & 27.60 & 0.825 & 0.215 & 1335.0 \\
    $\Delta$ (full--baseline) & -0.01 & -0.001 & +0.002 & -150.0 \\
    + SkipGS w/o budget & 27.23 & 0.804 & 0.249 & 1026.0 \\
    $\Delta$ (w/o--baseline) & -0.38 & -0.022 & +0.036 & -459.0 \\
    \midrule
    Taming 3DGS (baseline) & 27.94 & 0.822 & 0.207 & 757.0 \\
    + SkipGS (full) & 27.92 & 0.822 & 0.209 & 392.0 \\
    $\Delta$ (full--baseline) & -0.02 & 0.000 & +0.002 & -365.0 \\
    + SkipGS w/o budget & 27.27 & 0.795 & 0.260 & 161.0 \\
    $\Delta$ (w/o--baseline) & -0.67 & -0.027 & +0.053 & -596.0 \\
    \bottomrule
  \end{tabular}
\end{table}

\paragraph{Effect of backward budget control.}
Across both baselines, the full SkipGS pipeline achieves substantial post-densification speedups while preserving quality: PSNR drops by only 0.01\,dB on GaussianSpa and 0.02\,dB on Taming 3DGS, with SSIM and LPIPS nearly unchanged.
Removing budget control yields further time savings (e.g., $T_{\text{post}}$: $757$\,s $\rightarrow$ $161$\,s for Taming 3DGS), but at severe quality cost: PSNR drops by 0.67\,dB and 0.38\,dB on Taming 3DGS and GaussianSpa, respectively, with large degradation in SSIM and LPIPS.
Without the budget mechanism, the gating becomes overly aggressive; too many backward passes are skipped, starving the optimizer of gradient signal.
The budget controller prevents this by forcing backward execution when the cumulative ratio falls below $\rho_{\min}$ (Eq.~\ref{eq:rhomincalib}), recovering nearly all quality while retaining the majority of the speedup.
This confirms that budget control is essential for balancing acceleration and quality preservation across different base methods.

\section{Conclusion}
We presented \emph{SkipGS}, a view-adaptive backward gating mechanism for the post-densification phase of 3D Gaussian Splatting~\cite{kerbl3Dgaussians}.
By tracking per-view loss statistics and selectively skipping backward passes when a view's loss is consistent with its recent baseline, SkipGS reduces redundant computation where the backward pass dominates runtime.
A warmup-calibrated minimum backward budget preserves stable optimization.
Experiments on Mip-NeRF\,360~\cite{barron2022mipnerf360}, Deep Blending~\cite{hedman2018deepblending}, and Tanks\&Temples~\cite{Knapitsch2017tanks} show consistent wall-clock reductions with comparable reconstruction quality.
Last but not least, our experiments are limited to static scenes, and we leave the evaluation of SkipGS on dynamic scenes and models~\cite{Wu_2024_CVPR} for future work.

\section*{Acknowledgment}
Supported by the Intelligence Advanced Research Projects Activity (IARPA) via Department of Interior/ Interior Business Center (DOI/IBC) contract number 140D0423C0076. The U.S. Government is authorized to reproduce and distribute reprints for Governmental purposes notwithstanding any copyright annotation thereon. Disclaimer: The views and conclusions contained herein are those of the authors and should not be interpreted as necessarily representing the official policies or endorsements, either expressed or implied, of IARPA, DOI/IBC, or the U.S. Government.

%
%
\bibliographystyle{splncs04}
\bibliography{main}
\end{document}